\documentclass[11pt]{article}

\usepackage[final]{acl}
\usepackage{times}
\usepackage{latexsym}
\usepackage[T1]{fontenc}
\usepackage[utf8]{inputenc}
\usepackage{microtype}
\usepackage{inconsolata}
\usepackage{graphicx}
\usepackage{booktabs}
\usepackage{multirow}
\usepackage{colortbl}
\usepackage{xcolor}
\usepackage{array}
\usepackage{adjustbox}
\usepackage{makecell}
\usepackage{listings}
\usepackage{threeparttable}
\usepackage[dvipsnames]{xcolor}
\usepackage{xspace}

\newcommand{\DESQ}{%
  \textbf{D}%
  \textbf{e}%
  \textbf{S}%
  \textbf{Q}%
  \xspace%  ← ajoute automatiquement un espace si nécessaire
}

\title{\DESQ: \underline{De}composition-based
\underline{S}PARQL \underline{Q}uery
Generation}

\author{
  \begin{minipage}{\textwidth}
    \centering
    \begin{minipage}{0.45\textwidth}
      \centering
      \textbf{Papa Abdou Karim Karou Diallo$^{\dagger,\diamond,\S}$}
    \end{minipage}
    \begin{minipage}{0.45\textwidth}
      \centering
      \textbf{Aditya Sharma $^{\dagger,\diamond,\S}$}
    \end{minipage}
    \\[0.3em]
    \begin{minipage}{0.45\textwidth}
      \centering
      \textbf{Neshat Elhami Fard$^{\dagger,\diamond,\S}$}
    \end{minipage}
    \begin{minipage}{0.45\textwidth}
      \centering
      \textbf{Amal Zouaq$^{\dagger,\diamond,\S}$}
    \end{minipage}
    \\[0.5em]
    \normalfont
    $^\dagger$LAMA-WeST \quad
    $^\diamond$Mila -- Quebec AI Institute \quad
    $^\S$Polytechnique Montr\'eal \\[0.2em]
    {\small Correspondence: \href{mailto:diallokarou28@polymtl.ca}{\texttt{karou.diallo@mila.quebec}}}
  \end{minipage}
}

\begin{document}
\maketitle
\begin{abstract}
Dominant approaches to Knowledge Base Question Answering (KBQA) fall into two categories. First is the generation of a formal query that suffers from brittleness and limited explainability, and the second is direct answer retrieval through KB exploration that is computationally costly and prone to hallucination. To combine the strengths of both paradigms while mitigating their respective weaknesses, we introduce \DESQ (Decomposition-based SPARQL Query Generation), a KB-agnostic framework that operates in three stages. First, it decomposes complex questions into \texttt{Atomic Constraints (ACs)} that mirror the relational structure of the underlying KB. Second, it generates a two-part structured output: (a) \textbf{Mapping} of each \texttt{AC} to its corresponding SPARQL Fragment, using standardized variable and URIs placeholders, and (b) \textbf{URIs Grounding} block describing each placeholder. Third, it assembles these fragments into a complete SPARQL query. \DESQ surpasses state-of-the-art approaches on four out of five major benchmarks and demonstrates superior robustness to lexical variation.  Beyond performance gains, our framework greatly simplifies evaluation by eliminating the need for a live KB endpoint, and its structured output enables fine-grained error analysis, allowing more targeted interventions for improvement.

\begin{figure}[t]
  \includegraphics[width=\columnwidth]{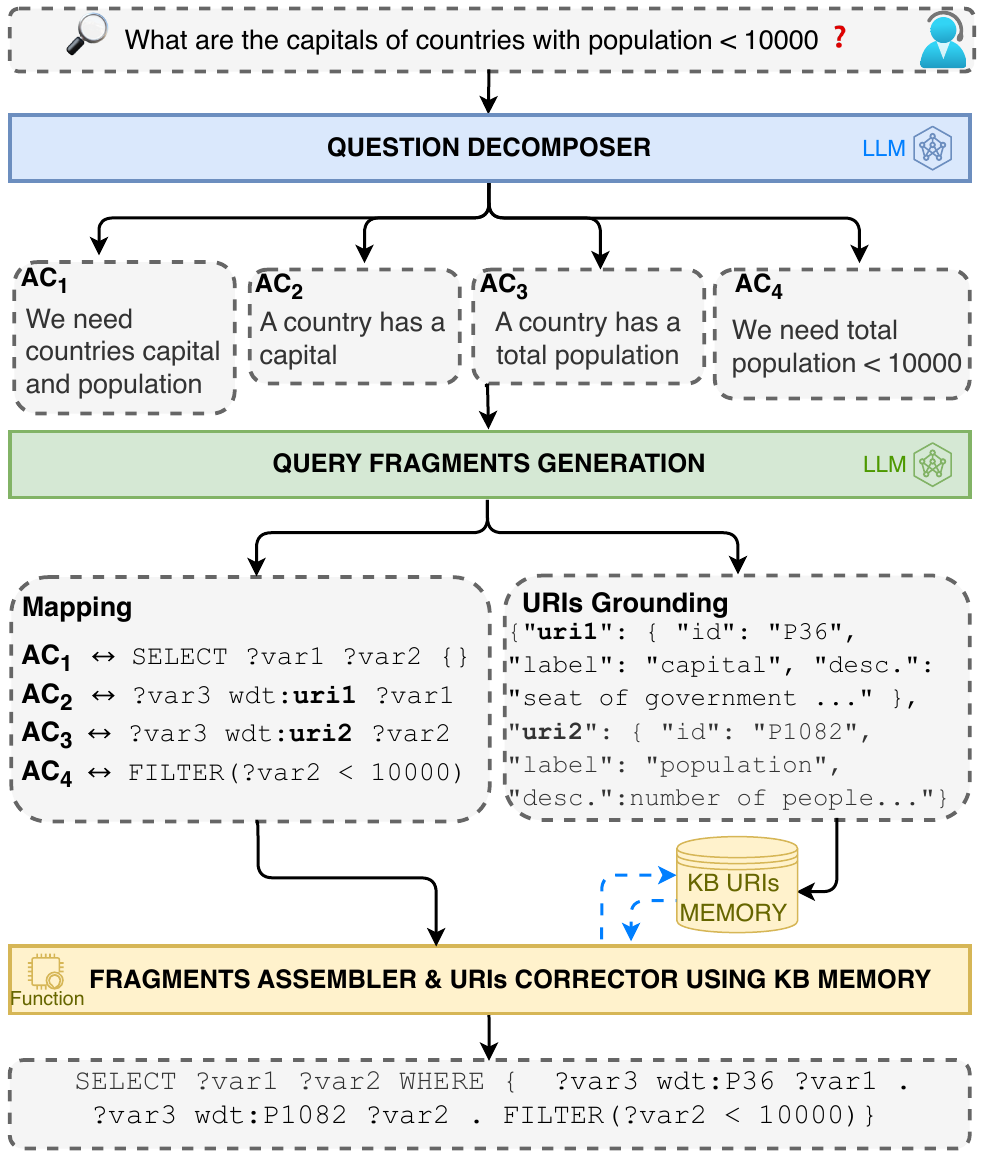}
  \caption{\DESQ's architecture.}
  \label{fig:ragex-pipeline}
\end{figure}
\end{abstract}

%%%%%%%%%%%%%%%%%%%%%%%%%%%%%%%%%%%%%%%%%%%%%%%%%%%%%%%%%%%%%%%%%
\section{Introduction}\label{sec:intro}
SPARQL and SQL have established themselves as the standard interface for querying RDF-based Knowledge Bases (KBs) \citep{gu2022knowledge} and Relational Databases (RDBs) \citep{liu2025survey}, enabling precise and reproducible retrieval over large-scale structured repositories. Yet composing correct SPARQL/SQL queries requires both familiarity with the query language and detailed knowledge of the underlying schema, creating a significant barrier for non-expert users\citep{hong2025next}. This has motivated growing interest in systems that automatically translate natural language questions into executable SPARQL/SQL queries \citep{diallo2024comprehensive, reyd2023assessing, hong2025next, li2024codes}.
State-of-the-art systems for this task fall into two dominant paradigms. The first generates structured queries directly from natural-language questions \citep{sharma2026reducing, banerjee2022modern, emonet2024llm}. While appealing in its directness, this approach struggles with explainability and brittle alignment between unstructured language and the highly structured identifiers of KBs, which often leads to malformed queries and incorrect answers. The second bypasses query generation altogether by directly exploring the KB to produce answers \citep{shavarani2024entity, alawwad2024enhancing, muennighoff2022sgpt}. Although this avoids the syntactic constraints of formal query languages, it is prone to hallucinations and entails expensive graph traversal that undermines both trust and efficiency \citep{liu2024spinach, li2023few, sun2025search}.
Generating a logical form that can retrieve information from a trustworthy, controlled, and continuously updated KB remains the more principled approach, as it preserves full auditability and supports exact, reproducible answers \citep{hong2025next}. However, expecting a model to produce a correct and executable SPARQL query in a single step is overly optimistic. Achieving this requires satisfying several interdependent prerequisites simultaneously: the model must capture the semantic structure of complex questions and align it with the schema of the target KB \citep{diallo2025frase, zahera2024generating, d2025investigating}; it must correctly identify and disambiguate entities and relations despite significant lexical variation between the question and the KB schema \citep{cao2026out, qi2024enhancing, polonuer2026autonomous}; and it must faithfully encode every condition and constraint while preserving focus on the target information to be retrieved. The difficulty compounds further when questions involve multi-hop reasoning, nested conditions, or aggregations \citep{alekseev2025benefits, mountantonakis2025generating, piao2026lite}.
These challenges motivate a decomposition-based approach. Breaking the generation process into modular steps makes the overall problem more tractable, enables fine-grained error analysis by isolating which step failed, and decouples linguistic understanding from KB-specific identifier resolution, two sources of error that are frequently conflated in end-to-end approaches \citep{diallo2024comprehensive}.
To this end, we introduce \DESQ, a KB-agnostic framework that combines the strengths of both paradigms while mitigating their respective weaknesses. As illustrated in Figure~\ref{fig:ragex-pipeline}, \DESQ operates in three stages. First, it decomposes natural language questions into \texttt{Atomic Constraints (ACs)} that reflect the relational structure of the underlying KB. It then generates a two-part structured output: (a) \textbf{Mapping} from each AC to its corresponding SPARQL Fragment, using a unified standardized naming convention for variables and URIs placeholders, and (b) \textbf{URIs Grounding} block providing the label and description of each URI placeholder. Finally, a deterministic assembler resolves these placeholders against a non-parametric KB memory and produces a complete, executable SPARQL query.
This work investigates three research questions. \\ 
\textbf{RQ1} How does a decomposition-based approach compare to direct SPARQL query generation? \\
\textbf{RQ2} How do the type, scale, and architecture of the underlying LLM affect \DESQ's performance? \\
\textbf{RQ3} How does \DESQ compare to previous state-of-the-art KBQA approaches across benchmarks?

The main contributions of this work are as follows: (1) we introduce \DESQ, a novel KB-agnostic KBQA framework whose design decouples linguistic reasoning from factual grounding through a structured decomposition-to-query pipeline. (2) We propose an efficient post-hoc URI correction strategy that leverages the generated label and description metadata to recover incorrect KB identifiers in the final SPARQL query, while selectively preserving URIs that were already predicted correctly. (3) \DESQ achieves state-of-the-art performance on four major benchmarks, namely LC-QuAD 2.0 \citep{dubey2019lcquad2}, WebQSP \citep{yih2016value}, ComplexWebQuestions \citep{talmor2018web}, and QALD-10 \citep{usbeck2024qald10} proving stronger generalization compared to existing approaches. (4) The structured output of \DESQ eliminates the need for a live KB endpoint during evaluation, avoiding the overestimation pitfall of execution-based F1 where empty-result query pairs are counted as correct regardless of semantic equivalence, while also enabling fine-grained error analysis at the level of individual steps.

%%%%%%%%%%%%%%%%%%%%%%%%%%%%%%%%%%%%%%%%%%%%%%%%%%%%%%%%%%%%%%%%%
\section{Related Work}\label{sec:related-works}
\subsection{SPARQL Query Generation} \label{sec:related-works-sparql-generation}
% 1st: Intermediate Query Generation and handling different ontologies schemas
Within the query generation paradigm, methods differ in how they handle the heterogeneity and complexity of the underlying KB. Several works address the diversity of available knowledge sources by merging datasets from different ontologies \citep{vollmers2024uniq}, aligning ontological representations, integrating federated knowledge graphs \citep{emonet2024llm}, or translating between ontologies \citep{bartels2025automating}. Others collect evidence from multiple sources before generating a query \citep{lv2020graph, vollmers2024uniq} or generate another logical form and then convert it to another target one \citep{luo2023chatkbqa}. Some methods enable LLMs to incorporate new information from external memory at inference time to reduce reliance on parametric knowledge \citep{zhang2024learn, sharma2026reducing}.
% 2nd: Direct Answer extraction family of methods
A complementary line of work bypasses formal query generation in favor of direct answer extraction. Some approaches gather relevant rationales incrementally \citep{li2023chain, zhao2023verify, wei2022chain} or plan an explicit sequence of reasoning steps prior to producing the final answer \citep{sun2023think}. Others leverage the native reasoning capabilities of LLMs directly \citep{jiang2023structgpt, gu2023don}, or perform graph traversal from an anchor entity until the answer is reached \citep{sun2025search}.

% Limitations of these approaches
% Existing approaches in both families share notable limitations that \DESQ is designed to address. Methods that generate queries end-to-end leave the intermediate decomposition unconstrained, with no guarantee that the resulting steps align with how information is organized in the target KB \citep{zhang2025kag}. Exploration-based methods require a correctly identified starting entity, which precludes answering certain question types present in standard benchmarks \citep{sun2025search}. Finally, approaches that plan reasoning steps freely do not enforce the correspondence between step granularity and the schema of the knowledge source \citep{sun2023think}.
%---------------------------------------------------------
\subsection{Complex Problem Solving Through Decomposition}\label{sec:related-works-decomposition-in-complex-problem-solving}
% Families of methods
Divide-and-conquer strategies have long been central to complex question answering, and their combination with LLMs has given rise to a rich family of decomposition-based methods \citep{hu2025divide, yao2023tree}. A first line of work focuses on eliciting structured reasoning through prompting, where intermediate steps are generated before producing a final answer \citep{wei2022chain}. Within this family, several approaches interleave reasoning with iterative retrieval over external sources, progressively gathering supporting evidence across multiple hops rather than retrieving all context upfront \citep{trivedi2023interleaving, sun2023think}. More recent contributions go further by training LLMs to autonomously decide when and how to invoke external search within a reasoning trace, using reinforcement learning to optimize the interplay between retrieval and reasoning \citep{li2025search, chen2025learning}.
A closely related family integrates semantic parsing with explicit structural decomposition, where questions are converted into logical or formal representations and then broken down into sub-queries whose dependencies are explicitly modeled \citep{liang2024kag, zhang2025kag}. Methods in this family differ in how they represent and resolve sub-problems, with some using a dual natural language and logical form representation to route each sub-problem to either retrieval or reasoning.
% % Limitations of these Families of methods
% While these approaches demonstrate the value of structured decomposition for logical coherence and retrieval precision, their decomposition granularity is not constrained to match the schema of the underlying knowledge source: sub-problems are generated freely by the model, without explicit alignment to how facts are stored in the KB. 

% Only one paragraph summarizing the limitations of approaches evoked in the two pervious sections and explaining how DeSQ differs from them
\DESQ differs fundamentally from both the SPARQL generation approaches (Section~\ref{sec:related-works-sparql-generation}) and the decomposition-based methods (Section~\ref{sec:related-works-decomposition-in-complex-problem-solving}) discussed above. Rather than treating query structure and KB-specific identifiers as a single entangled generation target, \DESQ explicitly supervises the model to distinguish between the two, learning linguistic query structure independently from factual KB grounding.
Second, instead of leaving decomposition granularity to the model's discretion, it explicitly constrains each decomposition step to correspond to a single relational triple in the KB schema, ensuring a direct structural correspondence between decomposition steps and the source of factual information. This stands in contrast to KAG \citep{liang2024kag}, whose decomposition targets rationale generation rather than schema-aligned query construction. Furthermore, unlike exploration-based methods \citep{sun2025search}, \DESQ does not require a starting entity and can handle the full range of question types found in standard KBQA benchmarks. To the best of our knowledge, such a granularity-aligned decomposition strategy has not previously been explored for SPARQL query generation or question answering.

%%%%%%%%%%%%%%%%%%%%%%%%%%%%%%%%%%%%%%%%%%%%%%%%%%%%%%%%%%%%%%%%%
\section{Methodology} \label{sec:methodology}
\subsection{Question decomposition} \label{sec:qst-decomposition}
We automatically construct gold decompositions of questions from the train set of each benchmark by extracting the list of fragments (\texttt{SELECT/FILTER} clause, Graph Patterns, and modifiers) from each associated gold SPARQL query and verbalizing each fragment into an \texttt{Atomic Constraint (AC)} using GPT-OSS 20B \citep{agarwal2025gpt} in inference mode with reasoning level set to \texttt{low}, ensuring structural alignment with the underlying KB's RDF format. 
For instance, given the question \textit{``What are the capitals of countries with population $<$ 10000 ?''} and its gold query \texttt{SELECT ?var1 ?var2 WHERE \{?var3 wdt:P36 ?var1 . ?var3 wdt:P1082 ?var2 . FILTER(?var2 $<$ 10000)\}}, four \texttt{ACs} are obtained from the verbalization of SPARQL Fragments. The fragment \texttt{SELECT ?var1 ?var2 \{\}} clause, which serves as the SPARQL query skeleton and whose verbalization captures the overall intent of the question, is mapped to \texttt{AC$_1$} \textit{``We need countries capital and population''}; the triple \texttt{?var3 wdt:P36 ?var1} into \texttt{AC$_2$} \textit{``A country has a capital''}; the triple \texttt{?var3 wdt:P1082 ?var2} into \texttt{AC$_3$} \textit{``A country has a total population''}; and the filter expression \texttt{FILTER(?var2 $<$ 10000)} into \texttt{AC$_4$} \textit{``We need total population $<$ 10000''}. This verbalization process is applied uniformly across all training examples. Importantly, at inference time, the model receives only the question and is expected to generate the corresponding \texttt{ACs} autonomously.

\subsection{Query Fragments Generation and URIs Grounding} \label{sec:query-fragment-generation}
From a question augmented with the decomposition (set of \texttt{ACs}) obtained at the previous step (Section~\ref{sec:qst-decomposition}), we generate two parts: (1) the \textbf{Mapping} between each \texttt{AC} to its SPARQL Fragment and the (2) \textbf{URIs Grounding}. This second part is represented as a list of dictionary format with ``id'', ``label'' and ``description'' as keys. For instance, \texttt{uri1} would be associated with \texttt{{``id'': ``P36'', ``label'': ``capital'', ``desc.'': ``seat of government of a country''}}. This is motivated by the fact that we already know the models tend to hallucinate a lot for these opaques identifier (IDs) \citep{diallo2024comprehensive} but we can later use the generated metadata to correct URIs IDs using the process defined in Section~\ref{sec:uris-correction}.

\subsection{Non-Parametric KB Memory} \label{sec:kb-memory}
To support URIs resolution without relying on a live KB endpoint, we build a non-parametric KB memory as a dense index over the KB's entities and relations. Each entry in the index stores the official identifier, label, and description of a KB element, encoded using BGE-Large\footnote{\url{https://huggingface.co/BAAI/bge-large-en}} as the embedding model. This memory serves as the backbone of the URIs correction mechanism, enabling efficient retrieval of candidate identifiers at inference time given a generated label and description.

% \subsection{Fragment Assembling \& URIs Correction Using KB Memory} \label{sec:fragments-assembling}
% This final step of our architecture is needed to get the final query from the fragments that are generated from the \texttt{ACs}. This process involve a deterministic algorithm and doesn't require any training or use of LLM. The URIs correction mechanism that is done in parallel aims to reduce the hallucinations and is explained in more details in Section~\ref{sec:uris-correction}.

\subsection{Fragment Assembling \& URIs Correction} \label{sec:fragments-assembling}
This final step assembles the generated SPARQL Fragments into a complete executable query through a deterministic algorithm that requires no additional training or LLM inference. In parallel, the URIs correction mechanism resolves predicted URI placeholders against the KB memory described in Section~\ref{sec:kb-memory}. The key insight behind this mechanism is that while the model is trained to predict both the KB identifier (e.g., \texttt{P36}), its human-readable label (e.g., \textit{``capital''}), and a natural language description (e.g., \textit{``seat of government of a country''}), the identifier alone is opaque and prone to hallucination. The label and description, however, are grounded in the natural language of the question and are therefore significantly easier for the model to predict correctly, and are thus used as semantic anchors to guide the correction of incorrect identifiers. This design allows the model to leverage its linguistic competence to guide factual KB grounding, while also enabling fine-grained error analysis by evaluating each field independently, as detailed in Section~\ref{sec:metrics}. The full correction strategy and assembling process are described in more detail in Section~\ref{sec:uris-correction}.

% \subsection{Training of the architecture}
% We've tested two version of the \DESQ's architecture: a module based one and an end-to-end version of it. The first requires to train each individual module to perform to designated sub-task while the second use only one training phase to perform the task and is the one we finally kept for \DESQ. %Details about results for these two architectural designs are provided in Appendix~\ref{comparison-RAGEX-e2e-vs-module-based}.

\subsection{Evaluation Metric} \label{sec:metrics}

Execution-based F1, as used by SOTA approaches, may overestimate true performance since query pairs with empty results are counted as correct matches. We instead adopt a triple-order-invariant Exact Match (EM) metric that checks whether each predicted triple pattern exactly matches its gold counterpart independently of order, operating on normalized variable names to abstract away superficial naming differences. The structured nature of \DESQ's output allows EM to be applied at three complementary levels of granularity: (i) fragment-level, assessing \texttt{AC}-to-SPARQL fragment correctness independently of URI resolution, (ii) URI-level, measuring the accuracy of predicted KB identifiers in the metadata block, and (iii) final query, comparing the fully assembled predicted query against the gold. This multi-granular protocol enables fine-grained error diagnosis, separating failures in linguistic query structure from failures in factual KB grounding. Further details about the evaluation pipeline are provided in Appendix \ref{sec:more-detail-evaluation-pipeline}.

\paragraph{Fragment-level evaluation.} \label{sec:uris-correction} For each example, the generated output is parsed to extract the mapping between \texttt{ACs} and SPARQL Fragments. Fragments are preprocessed prior to comparison through a normalization pipeline that includes lowercasing, standardization of operator spacing, and normalization of bracket spacing inside nested FILTER expressions. EM is computed per fragment and averaged across all fragments in the test set. Since key alignment between gold and predicted outputs may fail in cases where the model rephrases \texttt{ACs}, we evaluate under two alignment strategies: key-based alignment, where each predicted fragment is matched to its gold counterpart by key identity, and position-based alignment, where each predicted fragment is matched to the gold fragment occupying the same position in the output sequence (detailed illustrated shown in Appendix). Both strategies yield approximately same results. In addition to EM, we report corpus-level BLEU score computed over the full output sequences as a softer measure of lexical overlap.

\paragraph{URI-level evaluation.} \label{sec:uri-level-eval} Each URI placeholder is associated in the model output with a predicted KB identifier (e.g., \texttt{Q657072} or \texttt{m.05g477}), a label, and a description. We evaluate each field independently using EM and report the average across all URIs slots in the test set. Since KB identifiers are opaque and thus inherently difficult to predict correctly, we complement direct prediction with a retrieval-based correction stage against the KB memory (Section~\ref{sec:kb-memory}). We evaluate seven retrieval strategies, the best of which, termed \textit{Round-trip}, flags a predicted identifier as incorrect if it is absent from the KB memory or if its official label does not match the model-generated label under exact string matching, in which case a new dense retrieval search is performed using the concatenation of the generated label and description; otherwise, the original prediction is preserved. The remaining strategies are described in Appendix~\ref{sec:appendix-other-retrieval-strategies}.

\subsection{Datasets} \label{sec:datasets}

We evaluate \DESQ on five KBQA benchmarks of increasing complexity, summarized in Table~\ref{tab:datasets}. \textbf{WebQSP} \citep{yih2016value} and \textbf{LC-QuAD 2.0} (LCQ2) \citep{dubey2019lcquad2} are the simplest, both limited to two hops, over Freebase and Wikidata respectively. \textbf{CWQ} \citep{talmor2018web} extends \textbf{WebQSP} to four hops with richer SPARQL patterns. \textbf{GrailQA} \citep{gu2021beyond} is the primary out-of-distribution benchmark, evaluating i.i.d., compositional, and zero-shot generalization over Freebase. \textbf{QALD-10} \citep{usbeck2024qald10} is the most challenging, featuring expert-crafted questions over Wikidata with very high SPARQL diversity and a very small training set.

\begin{table*}[t]
\centering
\resizebox{\textwidth}{!}{%
\begin{tabular}{lcccccc}
\toprule
\textbf{Dataset} & \textbf{Knowledge Base} & \textbf{Size} & \textbf{Max Hops} & \textbf{Generalization} & \textbf{SPARQL Diversity} & \textbf{SOTA Exact Match} \\
\midrule
WebQSP       & Freebase            & 4,737  & 2  & i.i.d.                  & Low      & 78\% \citep{luo2023chatkbqa}  \\
LCQ2  & Wikidata + DBpedia  & 30,000 & 2  & i.i.d.                  & Medium   & --  \\
CWQ          & Freebase            & 34,689 & 4  & i.i.d.                  & Medium   & 77\% \citep{luo2023chatkbqa}  \\
GrailQA      & Freebase            & 64,331 & 4  & i.i.d. + Compositional + Zero-shot & High  & 79\% \citep{gao2025beyond} \\
QALD-10      & Wikidata            & 806    & 3+ & i.i.d.                  & Very High & -- \\
\bottomrule
\end{tabular}%
}
\caption{Comparison of KBQA benchmark datasets, ordered by increasing complexity.}
\label{tab:datasets}
\end{table*}

\subsection{Models} \label{sec:models}
To evaluate \DESQ, we experiment with six open-weight large language models spanning a wide range of sizes and architectural designs. We include \textbf{Qwen3.5-0.8B} and \textbf{Qwen3.5-27B} \citep{qwen3.5}, two language models built on a hybrid architecture combining Gated Delta Networks with sparse Mixture-of-Experts, trained with reinforcement learning at scale and supporting over 201 languages. The 0.8B variant serves as a lightweight baseline suitable for fine-tuning, while the 27B variant represents a higher-capacity configuration from the same family. We also include \textbf{Gemma 4-31B} \footnote{https://deepmind.google/models/gemma/gemma-4/}, which belongs to the Gemma 4 model family and is a dense multimodal model targeting higher reasoning capacity. We also evaluate \textbf{Llama3 8B} \citep{grattafiori2024llama}, a general-purpose open decoder model that serves as a widely adopted mid-scale baseline. Finally, we include  \textbf{Phi-4} \citep{abdin2024phi}, a 14B parameters dense decoder-only Transformer trained on a carefully curated blend of synthetic datasets and high-quality filtered web content, with a particular emphasis on advanced reasoning capabilities. Together, these models cover a broad spectrum of parameter counts (from 0.8B to 31B), architectural choices (dense vs. MoE), and training philosophies, allowing for a comprehensive comparative analysis of their ability to generate syntactically and semantically correct SPARQL queries from natural language inputs.

%%%%%%%%%%%%%%%%%%%%%%%%%%%%%%%%%%%%%%%%%%%%%%%%%%%%%%%%%%%%%%%%%
\section{Results and Discussion} \label{sec:results-discussion}

\begin{figure*}
  \includegraphics[width=\linewidth]{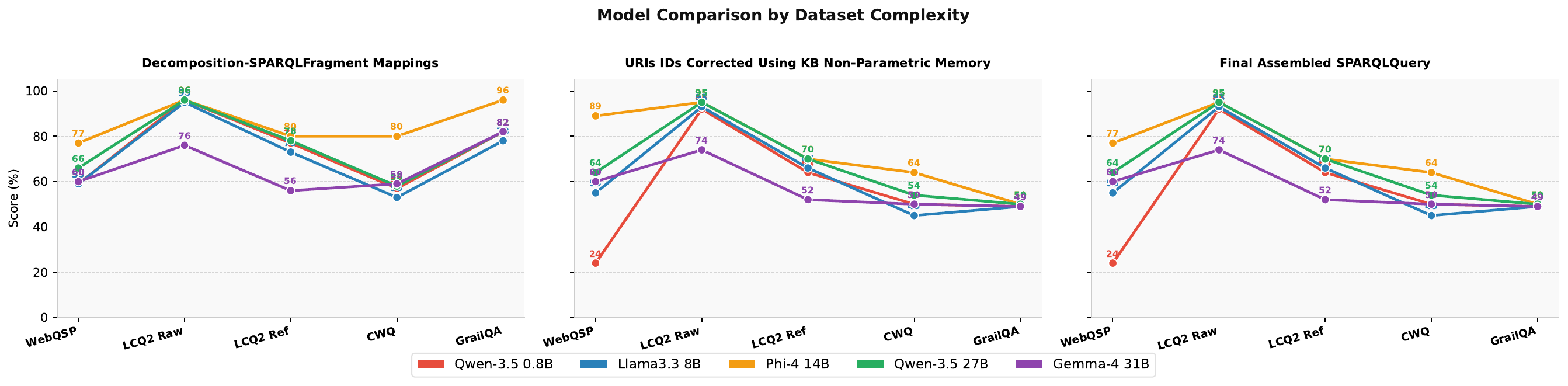}
  \caption{Models performances comparison across datasets on three evaluation criteria: SPARQL Fragment decomposition mapping, URIs Grounding identification, and final assembled query exact match.}
  \label{fig:spider-diagram-models-comparison-all-datasets}
\end{figure*}

% ------- GENERAL INSIGHTS REGARDING OUR APPROACH PERFORMANCE
\paragraph{General insights.}
As shown in Table~\ref{tab:models-comparison-all-datasets} and Figure~\ref{fig:spider-diagram-models-comparison-all-datasets}, all evaluated models achieve high Exact Match (EM) scores on the subtask of mapping \texttt{Atomic Constraints (ACs)} to SPARQL Fragments as depicted in the column `\textbf{Mapping}'. Regarding the ``\textbf{URIs Grounding}'' columns, models perform well on Label prediction but struggle more with ID prediction, which is expected given the opacity of KB identifiers. The ``\textbf{ID corrected}'' column, which reflects performance after the \textit{Round-trip} correction strategy (presented in Section~\ref{sec:uri-level-eval}), consistently improves over the raw ID prediction across all models, confirming the effectiveness of the post-hoc URI correction stage. This demonstrates their ability to understand the natural language question, infer the number of KB triples required to answer it, and correctly associate each \texttt{AC} with its corresponding SPARQL Fragment. More broadly, this confirms that \DESQ effectively enforces a clean separation between the structural dimension of the query language and its factual dimension, namely, the correct grounding of relations, classes, and entities to their KB-specific URIs.
When the performance on the fully assembled final query is lower (column `Exact Match Assembled Query'), it is typically because there is failure to effectively leverage the generated metadata to correct URIs IDs (column `URIs Grounding'). Indeed, this stage requires a comprehensive ontology and rich descriptions of KB entities to enable effective use of non-parametric memory, an aspect that is explicitly out of scope for this work. Nevertheless, this provides a strong guarantee: given access to a high-quality KB and ontology, very strong performance can be expected across all datasets and for the vast majority of models, including smaller ones such as Qwen3.5-0.8B that has competitive results compared with the Llama3.3 8B and Phi-4 14B despite the discrepancy in size.

\begin{table*}[!ht]
\centering
\adjustbox{max width=\textwidth}{%
\begin{tabular}{cc|cc|cc|cccc|c}
\toprule
%--- Header principal ---
\multirow{3}{*}{\textbf{Dataset}}
  & \multirow{3}{*}{\textbf{Models}}
  & \multicolumn{2}{c|}{\textbf{Global Output}}
  & \multicolumn{2}{c|}{\textbf{Mapping}}
  & \multicolumn{4}{c|}{\textbf{URIs Grounding}}
  & \multirow{3}{*}{\makecell{\textbf{Exact Match} \\ \textbf{Assembled} \\ \textbf{ Query \%}}} \\
\cmidrule{3-10}
& & EM & BLEU & EM & BLEU
  & \multicolumn{4}{c|}{Exact Match (EM)}
  & \\
& & & & & & ID & Label & Description & ID corrected & \\
\midrule
%--- LCQ2 Raw ---
\multirow{5}{*}{LCQ2 Raw}
  & Qwen-3.5 0.8B & 0 & 84 & 96 & 95 & 50 & 89 & 52 & 92 & 92 \\
\cline{2-11}
  & Llama3.3 8B   & 0 & 83 & 95 & 95 & 52 & 89 & 50 & 93 & 93 \\
\cline{2-11}
  & Phi-4 14B     & 0 & 86 & 96 & 95 & 70 & 92 & 64 & 95 & 95 \\
\cline{2-11}
  & Qwen-3.5 27B  & 0 & 85 & 96 & 96 & 57 & 91 & 56 & 95 & 95 \\
\cline{2-11}
  & Gemma-4 31B   & 0 & 66 & 76 & 75 & 40 & 71 & 41 & 74 & 74 \\
\midrule
\\
\midrule
%--- LCQ2 Ref ---
\multirow{5}{*}{LCQ2 Ref}
  & Qwen-3.5 0.8B & 0 & 71 & 77 & 83 & 31 & 59 & 35 & 64 & 64 \\
\cline{2-11}
  & Llama3.3 8B   & 0 & 70 & 73 & 81 & 32 & 61 & 34 & 66 & 66 \\
\cline{2-11}
  & Phi-4 14B     & 0 & 75 & 80 & 85 & 48 & 66 & 47 & 70 & 70 \\
\cline{2-11}
  & Qwen-3.5 27B  & 0 & 73 & 78 & 83 & 38 & 65 & 40 & 70 & 70 \\
\cline{2-11}
  & Gemma-4 31B   & 0 & 55 & 56 & 62 & 25 & 48 & 27 & 52 & 52 \\
\midrule
\\
\midrule
%--- WebQSP ---
\multirow{5}{*}{WebQSP}
  & Qwen-3.5 0.8B & 0 & 54 & 59 & 76 & 10 & 21 & 10 & 24 & 24 \\
\cline{2-11}
  & Llama3.3 8B   & 0 & 70 & 59 & 80 & 22 & 51 & 29 & 55 & 55 \\
\cline{2-11}
  & Phi-4 14B     & 0 & 89 & 80 & 95 & 72 & 87 & 78 & 89 & 80 \\
\cline{2-11}
  & Qwen-3.5 27B  & 0 & 76 & 66 & 86 & 39 & 60 & 40 & 64 & 64 \\
\cline{2-11}
  & Gemma-4 31B   & 0 & 73 & 60 & 80 & 31 & 58 & 40 & 60 & 60 \\
\midrule
\\
\midrule
%--- CWQ ---
\multirow{5}{*}{CWQ}
  & Qwen-3.5 0.8B & 0 & 64 & 57 & 63 & 42 & 54 & 48 & 59 & 59 \\
\cline{2-11}
  & Llama3.3 8B   & 0 & 63 & 53 & 61 & 35 & 48 & 42 & 55 & 55 \\
\cline{2-11}
  & Phi-4 14B     & 0 & 66 & 80 & 82 & 50 & 63 & 59 & 78 & 78 \\
\cline{2-11}
  & Qwen-3.5 27B  & 0 & 65 & 58 & 63 & 46 & 58 & 52 & 62 & 62 \\
\cline{2-11}
  & Gemma-4 31B   & 0 & 64 & 59 & 65 & 38 & 53 & 45 & 61 & 61 \\
\midrule
\\
\midrule
%--- GrailQA ---
\multirow{5}{*}{GrailQA}
  & Qwen-3.5 0.8B & 0 & 74 & 82 & 84 & 44 & 56 & 47 & 49 & 49 \\
\cline{2-11}
  & Llama3.3 8B   & 0 & 73 & 78 & 81 & 44 & 58 & 48 & 49 & 49 \\
\cline{2-11}
  & Phi-4 14B     & 0 & 80 & 96 & 97 & 29 & 54 & 26 & 66 & 66 \\
\cline{2-11}
  & Qwen-3.5 27B  & 0 & 74 & 82 & 85 & 46 & 60 & 50 & 50 & 50 \\
\cline{2-11}
  & Gemma-4 31B   & 0 & 75 & 82 & 85 & 44 & 58 & 47 & 49 & 49 \\
\midrule
\\
\midrule
%--- QALD-10 ---
\multirow{5}{*}{QALD-10}
  & Qwen-3.5 0.8B & 0 & 70 & 41 & 57 & 24 & 51 & 19 & 58 & 41 \\
\cline{2-11}
  & Llama3.3 8B   & 0 & 47 & 30 & 49 & 8 & 24 & 1 & 28 & 28 \\
\cline{2-11}
  & Phi-4 14B     & 0 & 86 & 66 & 73 & 59 & 74 & 50 & 78 & 66 \\
\cline{2-11}
  & Qwen-3.5 27B  & 0 & 75 & 58 & 67 & 39 & 64 & 25 & 70 & 58 \\
\cline{2-11}
  & Gemma-4 31B   & 0 & 2 & 6 & 0 & 2 & 0 & 1 & 2 & 2 \\
\bottomrule
\end{tabular}
}
\caption{Performances for all configurations of datasets and models}
\label{tab:models-comparison-all-datasets}
\end{table*}

\begin{table}
\centering
\begin{threeparttable}
\resizebox{\columnwidth}{!}{%
\begin{tabular}{lcc}
\toprule
\textbf{Datasets} 
  & \makecell{\textbf{Previous SOTA} \\ \textbf{Exact Match (\%)}} 
  & \makecell{\DESQ \\ \textbf{Exact Match (\%)}} \\
\midrule
LCQ2 Raw   & - & \textbf{95} \\

LCQ2 Ref   & - & \textbf{70} \\
\hline
WebQSP  & \makecell{77 \\ \citep{luo2023chatkbqa}} & \textbf{80} \\
\hline
CWQ     & \makecell{76 \\ \citep{luo2023chatkbqa}} & \textbf{78} \\
\hline
GrailQA & \makecell{\textbf{79} \\ \citep{gao2025beyond}} & 66 \\
\hline
QALD10  & - & \textbf{66} \\
\bottomrule
\end{tabular}
}
\begin{tablenotes}
\footnotesize
\item Our results are all obtained with Phi-4.
\end{tablenotes}
\caption{\DESQ performances vs. previous SOTA.}
\label{tab:DeSQ-vs-SOTA}
\end{threeparttable}
\end{table}

% ------- ANSWER TO RQ1 - How does a decomposition-based approach compare to direct SPARQL query generation? 
\paragraph{\DESQ vs.\ direct SPARQL query generation.} Directly fine-tuning models to generate SPARQL queries from questions proves highly ineffective when evaluation relies on Exact Match (EM) metric used in \DESQ. Specifically, when fine-tuning Phi-4 14B and Llama 3.3 8B under this direct generation paradigm, the EM score is very low across all datasets, which is expected given that even minor surface differences in variable naming, URI formatting, or whitespace between the predicted and gold queries result in a mismatch under strict exact matching. While execution-based F1 scores, which require a live SPARQL endpoint and are inherently more lenient than EM since they operate on the execution results of queries rather than verifying query structure, yield higher results (63\% for Phi-4 14B and 46\% for Llama 3.3 8B) \citep{diallo2025frase}, they remain substantially below the performance achieved by \DESQ. These results collectively underscore the soundness of our decomposition-based approach and its two-part structured output as shown in Section~\ref{sec:query-fragment-generation}.

% ------- ANSWER TO RQ2 - How do the type, scale, and architecture of the underlying LLM affect \DESQ's performance? 
\paragraph{Impact of models sizes/types and multi modality and reasoning.} Table~\ref{tab:models-comparison-all-datasets} shows performances for the different models architectures and sizes presented in Section~\ref{sec:models}. Despite being the biggest the model Gemma4 31B underperformed for a the majority of datasets. Dense models have outperform the counter-part MOE/hybrid (Qwen3.5 27B) and a bigger size of models doesn't lead to better performance. We also see that reasoning-based models such as Qwen3.3 and Gemma4 don't outperform the non-reasoning-based ones which makes sense since the format (style and length) of reasoning traces generated through \DESQ doesn't necessary align with the ones they were trained on as outlined by \citep{guo2025deepseek}.

\begin{figure*}
  \includegraphics[width=\linewidth]{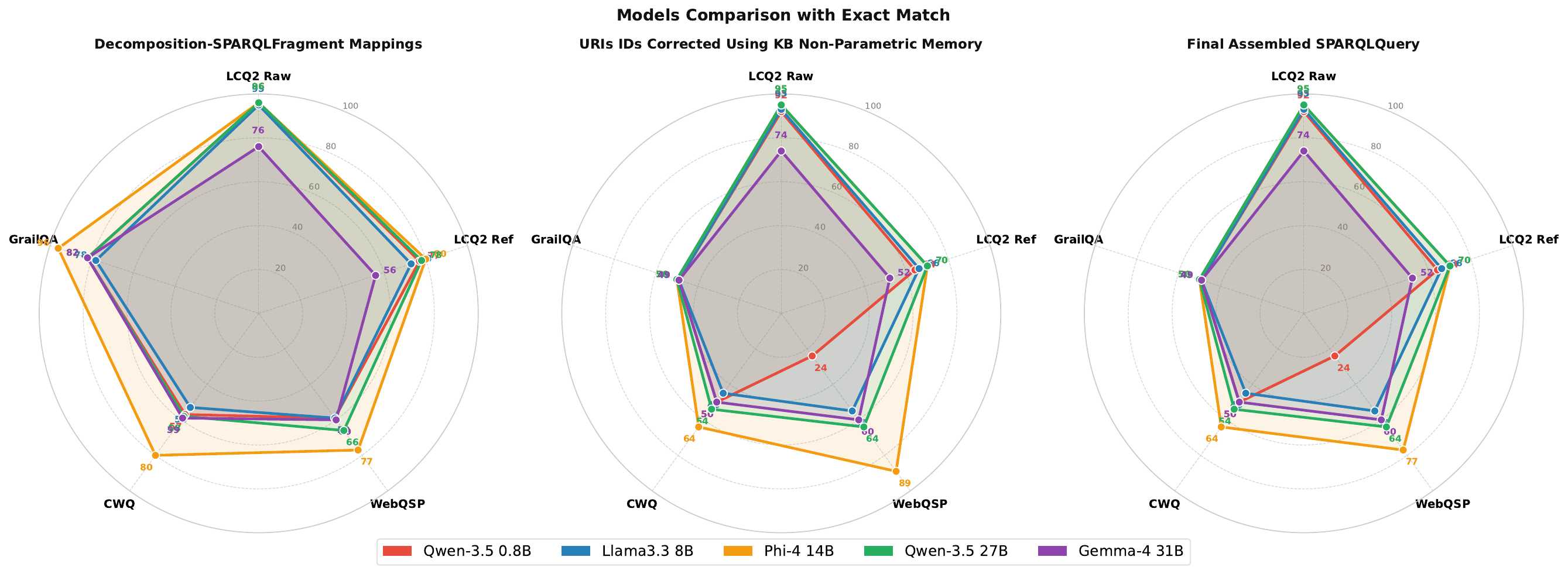}
  %\caption{Correlation between datasets complexity and performances in three criteria.}
  \caption{Performance of all evaluated models across datasets of increasing complexity (left to right).}%, measured at three complementary evaluation levels: Decomposition-to-SPARQL Fragment Mappings (left), URI identification after KB non-parametric memory correction (center), and Final Assembled SPARQL Query (right).}
  \label{fig:correlation-complexity-performance}
\end{figure*}

\paragraph{Correlation between performances and dataset complexity.} 
As shown in Figure~\ref{fig:correlation-complexity-performance}, the Decomposition-to-SPARQL Fragment Mappings panel reveals that models maintain relatively stable performance across datasets, indicating that the linguistic decomposition task is largely robust to dataset complexity. In contrast, the URI identification panel exhibits sharper degradation across datasets. The Final Assembled Query panel closely mirrors the URI identification panel, confirming that URI resolution is the primary performance bottleneck, and that this bottleneck is largely attributable to incomplete or poorly described KB elements that hinder effective non-parametric memory retrieval. All models converge toward a similar performance floor on GrailQA (49--50\%) regardless of size or architecture, pointing to a shared ceiling imposed by the limited ontological coverage of that benchmark. Throughout all panels, Phi-4 14B demonstrates the strongest resilience, while Gemma-4 31B consistently underperforms relative to its parameter count.

\paragraph{Case of QALD10.} QALD-10's high structural complexity combined with its very small training set makes it insufficient for the model to reliably learn our structured output format. We therefore apply continual learning from weights pre-trained on LCQ2, which shares the same underlying KB (Wikidata). This transfer strategy yields reasonable performance with Phi-4, though other evaluated models did not benefit from it to the same extent.

% ------- ANSWER TO RQ3 - How does \DESQ compare to previous state-of-the-art KBQA approaches across benchmarks?
\paragraph{\DESQ vs. previous SOTA.} Table~\ref{tab:DeSQ-vs-SOTA} reports the EM performance of \DESQ against previous SOTA approaches across all evaluated benchmarks. For LCQ2, where no previous EM-based comparison is available, the closest reference points are execution-based F1 scores of 94\% \citep{diallo2024comprehensive} and 67\% \citep{diallo2025frase} on the Raw and Reformulated splits respectively. \DESQ achieves 95\% and 70\% on the same splits under the stricter EM metric. %, the lower score on the Reformulated split being expected given its greater lexical and structural variation. 
\DESQ achieves consistent improvements on WebQSP and  CWQ compared with ChatKBQA approach \citep{luo2023chatkbqa}. Their evaluation treats URI retrieval as a black box, leaving their reported 27.2\% entity retrieval error rate undiagnosed at the intermediate level. They implicitly assume the LLM generates labels close enough to Freebase surface forms for successful retrieval, a fragile assumption our analysis shows fails on average in 39\% of evaluations on comparable data. Finally, since Answer EM passing does not imply logical form correctness, their F1 scores systematically overestimate true retrieval robustness, whereas our metrics provide verifiable URI-level grounding independent of downstream execution artifacts.
On QALD-10, for which no prior EM score was reported, we establish a first strong baseline at 66\%.  The only benchmark on which \DESQ does not reach the state of the art is GrailQA, where it scores 66\% against the 79\% reported by \citep{gao2025beyond}. This gap is not attributable to a structural limitation of the framework but rather to the incompleteness of KB element descriptions available for this benchmark, which hinders the URI resolution stage. %Crucially, all \DESQ results are obtained with a single model, Phi-4 (14B), without any dataset-specific architectural adaptation, which underscores the generalization capability. 
Crucially, all \DESQ results are obtained using the same model, Phi-4 (14B), with a uniform architecture and training procedure across all benchmarks, with no dataset-specific modifications, which underscores the generalization capability of the framework.
% Conclusion of this Results section
A key contribution of this work that could generalize to other NLP tasks is the explicit decoupling of linguistic knowledge from factual knowledge. By training the model to produce a structured two-part output, namely (i) the mapping between question decompositions and SPARQL Fragments, and (ii) the metadata of the URIs referenced in those fragments, the model is guided to learn these two distinct paradigms separately. This separation is particularly beneficial because it allows fine-grained error analysis: one can independently assess where the model fails on the linguistic structure of the query versus where it fails on factual grounding, and apply targeted corrections accordingly.

%%%%%%%%%%%%%%%%%%%%%%%%%%%%%%%%%%%%%%%%%%%%%%%%%%%%%%%%%%%%%%%%%

\section{Conclusion} \label{sec:conclusion}
We present \DESQ, a KB-agnostic framework for KBQA that operates through three structured stages: question decomposition, generation of structured output with SPARQL Fragment mapping and URIs Grounding, and query assembly. Our investigation reveals that constraining decomposition granularity to match the relational structure of the underlying KB is a critical design choice that prior work has overlooked, one that simultaneously improves query correctness, enables fine-grained error diagnosis, and removes the dependency on a live KB endpoint at evaluation time. We further show that post-hoc URIs IDs correction driven by generated label and description can substantially recover incorrect KB identifiers without altering those already correctly predicted. Across four out of the five benchmarks, \DESQ surpasses SOTA approaches and generalizes more robustly under lexical variation.

%%%%%%%%%%%%%%%%%%%%%%%%%%%%%%%%%%%%%%%%%%%%%%%%%%%%%%%%%%%%%%%%%
% \newpage
\section{Limitations} \label{sec:limitations}
Our approach relies on the completeness of the KBs, as URIs Grounding requires well-defined entities, relations, and classes. However, KBs are often incomplete, which constrains the availability and quality of such descriptions. Consequently, the effectiveness of our approach is limited by KB incompleteness.

%%%%%%%%%%%%%%%%%%%%%%%%%%%%%%%%%%%%%%%%%%%%%%%%%%%%%%%%%%%%%%%%%
\section{Acknowledgments} \label{sec:acknowledgments}
This project was undertaken thanks to funding from IVADO\footnote{https://ivado.ca/} and the Canada First Research Excellence Fund. We also acknowledge the financial support of the NSERC Discovery Grant Program, which contributed in part to this research. The authors further gratefully recognize Compute Canada (Calcul Québec) for providing the computational resources that made this work possible. We also benefited from technical support provided by Mila (mila.quebec).

\bibliography{custom}

%%%%%%%%%%%%%%%%%%%%%%%%%%%%%%%%%%%%%%%%%%%%%%%%%%%%%%%%%%%%%%%%%
\appendix
\section{Further details about the evaluation pipeline} \label{sec:more-detail-evaluation-pipeline}
\begin{figure*}
  \includegraphics[width=\linewidth]{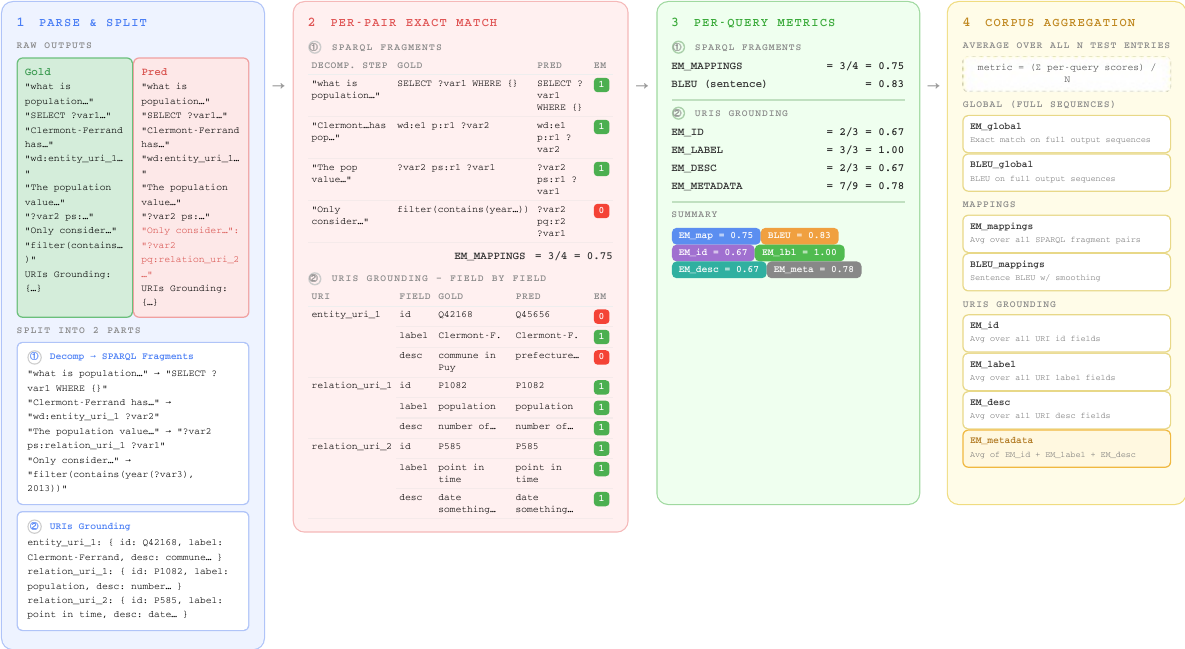}
  \caption{Order-Invariant Exact Match under normalized variables.}
  \label{fig:eval-pipeline}
\end{figure*}

Figure~\ref{fig:eval-pipeline} shows a concrete example of how the evaluation is performed for one entry and for the whole corpus.
Since the model is trained to generate a structured mapping where each \texttt{AC} serves as a key associated with its corresponding SPARQL Fragment, evaluation requires aligning predicted key-fragment pairs with their gold counterparts. However, the model may rephrase an \texttt{AC} slightly differently from the gold label, causing strict key matching to fail even when the associated SPARQL Fragment is correct. We therefore evaluate under two complementary strategies: \textbf{key-based alignment}, where predicted fragments are matched to gold counterparts by exact \texttt{AC} string matching, and \textbf{position-based alignment}, where fragments are matched by their sequential order in the output regardless of the \texttt{AC} label. The fact that both strategies yield approximately the same results indicates that key rephrasing is rare and that the model reliably preserves the ordering of decomposition steps.

\section{Additional Retrieval Strategies for URIs Correction using the External Non-Parametric KB Memory} \label{sec:appendix-other-retrieval-strategies}
\paragraph{Oracle correction.} serves as an upper bound or maximum improvement we could if we were just correcting the wrong URIs from the generated output, leaving the correct one unchanged. It applies retrieval using the label and description only for URIs whose predicted identifier differs from the gold identifier, leaving correctly predicted URIs untouched. This strategy yields an EM of 0.9061 and defines the ceiling achievable by any retrieval-based approach that perfectly identifies which URIs require correction.

\paragraph{Round-trip fuzzy.} follows the same detection principle but replaces exact string matching with a word-overlap criterion. The official label and the generated label are considered matching if their Jaccard word-level overlap reaches at least 50\%, accommodating minor paraphrasing or abbreviation in model-generated labels. This relaxed criterion reduces false detections on correct URIs, resulting in a small but consistent improvement over the exact variant.

\paragraph{Confidence score threshold.} takes a different approach by always performing a dense retrieval search using the generated label and description, regardless of whether the predicted identifier is suspected to be correct. The retrieved candidate is accepted as a replacement only if its cosine similarity score meets or exceeds a fixed threshold (set to 0.85 in our experiments). When the score falls below the threshold, the original predicted identifier is retained on the assumption that a low-confidence retrieval result is more likely to introduce an error than to correct one. The lower performance of this strategy relative to round-trip approaches suggests that unconditional retrieval frequently degrades already correct predictions when the similarity score criterion is not sufficiently discriminative.

\paragraph{Combined.} integrates the detection mechanism of round-trip exact with the confidence filtering of the score threshold strategy. A URI is first verified through round-trip exact matching; only when an error is detected is a dense retrieval search performed, and the retrieved candidate is accepted only if its similarity score meets the threshold. The two-stage gating is designed to avoid both unnecessary re-retrieval on correct URIs and unreliable corrections on uncertain retrieval results. Despite its conservative design, the combined strategy does not outperform the simpler round-trip approaches in our experiments, suggesting that the score threshold introduces missed corrections rather than preventing false ones at the threshold value used.

Tables~\ref{tab:detailed-eval-phi4-lcq2raw} and~\ref{tab:detailed-eval-phi4-lcq2ref} show the results for all these strategies (Strat. for short in the tables) used to correct URIs using the KB memory. Across all retrieval strategies, \textbf{round-trip fuzzy} achieves the best performance. 

\begin{table*}
\centering
\adjustbox{max width=\textwidth}{%
\begin{tabular}{cc|cc|cccccc}
\toprule
%--- Header principal ---
\multicolumn{2}{c|}{\textbf{Global Output}}
  & \multicolumn{2}{c|}{\textbf{Mappings}}
  & \multicolumn{6}{c}{\textbf{URIs Grounding}} \\
\midrule
%--- Sous-en-tête ligne 1 ---
Exact Match & BLEU & Exact Match & BLEU
  & \multicolumn{1}{c}{}
  & \multicolumn{4}{c}{Exact Match}
  & \multicolumn{1}{c}{} \\
%--- Sous-en-tête ligne 2 ---
& & & & & ID & Label & Description & AVG & \\
\midrule
%--- Données principales ---
0 & 84 & 96 & 95 & & 70 & 92 & 64 & 76 & \\
\midrule
%--- Titre section URIs correction ---
\multicolumn{1}{c}{} &
\multicolumn{1}{c}{} &
\multicolumn{8}{c}{\textbf{URIs Correction using the KB as External Memory}} \\
\midrule
%--- En-têtes des 8 stratégies ---
\multicolumn{1}{c}{} &
\multicolumn{1}{c}{} &
\multicolumn{1}{c}{\makecell{Strat. 1 \\ \textit{Label}}} &
\multicolumn{1}{c}{\makecell{Strat. 2 \\ \textit{Desc}}} &
\multicolumn{1}{c}{\makecell{Strat. 3 \\ \textit{Label+Desc}}} &
\multicolumn{1}{c}{\makecell{Strat. 4 \\ \textit{Look-up}}} &
\multicolumn{1}{c}{\makecell{\textcolor{gray}{Upper} \\ \textcolor{gray}{\textit{Bound}}}} &
\multicolumn{1}{c}{\makecell{Strat. 6 \\ \textit{roundtrip}}} &
\multicolumn{1}{c}{\makecell{Strat. 7 \\ \textit{rt\_fuzzy}}} &
\multicolumn{1}{c}{\makecell{Strat. 8 \\ \textit{thresh}}} \\
%--- Valeurs des 8 stratégies ---
\multicolumn{1}{c}{} &
\multicolumn{1}{c}{} &
\multicolumn{1}{c}{90} &
\multicolumn{1}{c}{79} &
\multicolumn{1}{c}{95} &
\multicolumn{1}{c}{93} &
\multicolumn{1}{c}{\textcolor{gray}{96}} &
\multicolumn{1}{c}{\textbf{95}} &
\multicolumn{1}{c}{94} &
\multicolumn{1}{c}{83} \\
\bottomrule
\end{tabular}
}
\caption{Detailed Evaluation of Phi-4's Performance on LCQ2 Raw}
\label{tab:detailed-eval-phi4-lcq2raw}
\end{table*}

\begin{table*}
\centering
\adjustbox{max width=\textwidth}{%
\begin{tabular}{cc|cc|cccccc}
\toprule
%--- Header principal ---
\multicolumn{2}{c|}{\textbf{Global Output}}
  & \multicolumn{2}{c|}{\textbf{Mappings}}
  & \multicolumn{6}{c}{\textbf{URIs Grounding}} \\
\midrule
%--- Sous-en-tête ligne 1 ---
Exact Match & BLEU & Exact Match & BLEU
  & \multicolumn{1}{c}{}
  & \multicolumn{4}{c}{Exact Match}
  & \multicolumn{1}{c}{} \\
%--- Sous-en-tête ligne 2 ---
& & & & & ID & Label & Description & AVG & \\
\midrule
%--- Données principales ---
0 & 74.77 & 80.25 & 85.20 & & 47.69 & 65.93 & 46.96 & 53.53 & \\
\midrule
%--- Titre section URIs correction ---
\multicolumn{1}{c}{} &
\multicolumn{1}{c}{} &
\multicolumn{8}{c}{\textbf{URIs Correction using the KB as External Memory}} \\
\midrule
%--- En-têtes des 8 stratégies ---
\multicolumn{1}{c}{} &
\multicolumn{1}{c}{} &
\multicolumn{1}{c}{\makecell{Strat. 1 \\ \textit{Label}}} &
\multicolumn{1}{c}{\makecell{Strat. 2 \\ \textit{Desc}}} &
\multicolumn{1}{c}{\makecell{Strat. 3 \\ \textit{Label+Desc}}} &
\multicolumn{1}{c}{\makecell{Strat. 4 \\ \textit{Look-up}}} &
\multicolumn{1}{c}{\makecell{\textcolor{gray}{Strat. 5} \\ \textcolor{gray}{\textit{Trg URIs}}}} &
\multicolumn{1}{c}{\makecell{Strat. 6 \\ \textit{roundtrip}}} &
\multicolumn{1}{c}{\makecell{Strat. 7 \\ \textit{rt\_fuzzy}}} &
\multicolumn{1}{c}{\makecell{Strat. 8 \\ \textit{thresh}}} \\
%--- Valeurs des 8 stratégies ---
\multicolumn{1}{c}{} &
\multicolumn{1}{c}{} &
\multicolumn{1}{c}{67.10} &
\multicolumn{1}{c}{57.31} &
\multicolumn{1}{c}{70.35} &
\multicolumn{1}{c}{65.82} &
\multicolumn{1}{c}{\textcolor{gray}{71.12}} &
\multicolumn{1}{c}{\textbf{70.45}} &
\multicolumn{1}{c}{69.64} &
\multicolumn{1}{c}{59.48} \\
\bottomrule
\end{tabular}
}
\caption{Detailed Evaluation of Phi-4's Performance on LCQ2 Ref Qst}
\label{tab:detailed-eval-phi4-lcq2ref}
\end{table*}

%%%%%%%%%%%%%%%%%%%%%%%%%%%%%%%%%%%%%%%%%
% DESQ generalization under lexical variation.
%%%%%%%%%%%%%%%%%%%%%%%%%%%%%%%%%%%%%%%%%

%--- TABLE 6: Train Raw Qst, Test Ref Qst ---
\begin{table*}
\centering
\adjustbox{max width=\textwidth}{%
\begin{tabular}{cc|cc|cccccc|c}
\toprule
\multicolumn{2}{c|}{\textbf{Global Output}}
  & \multicolumn{2}{c|}{\textbf{Mappings}}
  & \multicolumn{6}{c|}{\textbf{URIs Grounding}}
  & \multirow{3}{*}{\makecell{\textbf{Exact Match} \\ \textbf{Assembled} \\ \textbf{Query \%}}} \\
\cmidrule{1-10}
Exact Match & BLEU & Exact Match & BLEU
  & \multicolumn{1}{c}{}
  & \multicolumn{4}{c}{Exact Match}
  & \multicolumn{1}{c|}{} & \\
& & & & & ID & Label & Description & AVG & & \\
\midrule
0 & 81.86 & 89.65 & 91.42 & & 60.37 & 83.25 & 59.18 & 67.60 & & 87.20 \\
\midrule
\multicolumn{1}{c}{} &
\multicolumn{1}{c}{} &
\multicolumn{8}{c|}{\textbf{URIs Correction using the KB as External Memory}} & \\
\midrule
\multicolumn{1}{c}{} &
\multicolumn{1}{c}{} &
\multicolumn{1}{c}{\makecell{Strat. 1 \\ \textit{Label}}} &
\multicolumn{1}{c}{\makecell{Strat. 2 \\ \textit{Desc}}} &
\multicolumn{1}{c}{\makecell{Strat. 3 \\ \textit{Label+Desc}}} &
\multicolumn{1}{c}{\makecell{Strat. 4 \\ \textit{Look-up}}} &
\multicolumn{1}{c}{\makecell{\textcolor{gray}{Strat. 5} \\ \textcolor{gray}{\textit{Trg URIs}}}} &
\multicolumn{1}{c}{\makecell{Strat. 6 \\ \textit{roundtrip}}} &
\multicolumn{1}{c}{\makecell{Strat. 7 \\ \textit{rt\_fuzzy}}} &
\multicolumn{1}{c}{\makecell{Strat. 8 \\ \textit{thresh}}} & \\
\multicolumn{1}{c}{} &
\multicolumn{1}{c}{} &
\multicolumn{1}{c}{82.77} &
\multicolumn{1}{c}{72.20} &
\multicolumn{1}{c}{87.08} &
\multicolumn{1}{c}{83.44} &
\multicolumn{1}{c}{\textcolor{gray}{87.76}} &
\multicolumn{1}{c}{\textbf{87.20}} &
\multicolumn{1}{c}{86.09} &
\multicolumn{1}{c}{75.02} & \\
\bottomrule
\end{tabular}
}
\caption{Detailed Evaluation of Phi-4's Performance -- Train: Raw Qst, Test: Ref Qst}
\label{tab:detailed-eval-phi4-raw2ref}
\end{table*}

%--- TABLE 7: Train Ref Qst, Test Raw Qst ---
\begin{table*}
\centering
\adjustbox{max width=\textwidth}{%
\begin{tabular}{cc|cc|cccccc|c}
\toprule
\multicolumn{2}{c|}{\textbf{Global Output}}
  & \multicolumn{2}{c|}{\textbf{Mappings}}
  & \multicolumn{6}{c|}{\textbf{URIs Grounding}}
  & \multirow{3}{*}{\makecell{\textbf{Exact Match} \\ \textbf{Assembled} \\ \textbf{Query \%}}} \\
\cmidrule{1-10}
Exact Match & BLEU & Exact Match & BLEU
  & \multicolumn{1}{c}{}
  & \multicolumn{4}{c}{Exact Match}
  & \multicolumn{1}{c|}{} & \\
& & & & & ID & Label & Description & AVG & & \\
\midrule
0 & 61.38 & 50.76 & 61.51 & & 31.58 & 48.00 & 33.17 & 37.58 & & 50.76 \\
\midrule
\multicolumn{1}{c}{} &
\multicolumn{1}{c}{} &
\multicolumn{8}{c|}{\textbf{URIs Correction using the KB as External Memory}} & \\
\midrule
\multicolumn{1}{c}{} &
\multicolumn{1}{c}{} &
\multicolumn{1}{c}{\makecell{Strat. 1 \\ \textit{Label}}} &
\multicolumn{1}{c}{\makecell{Strat. 2 \\ \textit{Desc}}} &
\multicolumn{1}{c}{\makecell{Strat. 3 \\ \textit{Label+Desc}}} &
\multicolumn{1}{c}{\makecell{Strat. 4 \\ \textit{Look-up}}} &
\multicolumn{1}{c}{\makecell{\textcolor{gray}{Strat. 5} \\ \textcolor{gray}{\textit{Trg URIs}}}} &
\multicolumn{1}{c}{\makecell{Strat. 6 \\ \textit{roundtrip}}} &
\multicolumn{1}{c}{\makecell{Strat. 7 \\ \textit{rt\_fuzzy}}} &
\multicolumn{1}{c}{\makecell{Strat. 8 \\ \textit{thresh}}} & \\
\multicolumn{1}{c}{} &
\multicolumn{1}{c}{} &
\multicolumn{1}{c}{52.81} &
\multicolumn{1}{c}{42.68} &
\multicolumn{1}{c}{54.86} &
\multicolumn{1}{c}{48.04} &
\multicolumn{1}{c}{\textcolor{gray}{56.15}} &
\multicolumn{1}{c}{\textbf{54.92}} &
\multicolumn{1}{c}{54.62} &
\multicolumn{1}{c}{43.58} & \\
\bottomrule
\end{tabular}
}
\caption{Detailed Evaluation of Phi-4's Performance -- Train: Ref Qst, Test: Raw Qst}
\label{tab:detailed-eval-phi4-ref2raw}
\end{table*}

\section{\DESQ generalization under lexical variation} Tables~\ref{tab:detailed-eval-phi4-raw2ref} and~\ref{tab:detailed-eval-phi4-ref2raw} evaluate the cross-version generalization capacity of \DESQ under two contrasting train/test configurations. To this end, we leverage LCQ2, a dataset that natively provides two complementary formulations for each question: a template-based version (Raw Qst), constructed from structured patterns, and a naturally reformulated version (Ref Qst), written in free-form natural language. By training on one formulation and testing on the other, we can directly measure how well the model transfers across question styles without any additional supervision. These two tables evaluate cross-version generalization: how well a model trained on one question formulation (Raw or Ref) transfers to the other at test time.
The asymmetry between the two configurations is striking. When trained on Raw Qst and tested on Ref Qst (Table~\ref{tab:detailed-eval-phi4-raw2ref}), the model achieves a URI AVG of 67.60 and a best corrected score of 87.20 (Strat. 6), indicating a reasonable capacity to generalize from naturally phrased to reformulated questions. The Mappings EM of 89.65 further confirms that structural query generation transfers relatively well in this direction.
The reverse configuration, trained on Ref Qst, tested on Raw Qst (Table~\ref{tab:detailed-eval-phi4-ref2raw}), tells a very different story. The URI AVG collapses to 37.58, Mappings EM drops to 50.76, and even the best corrected URI score (Strat. 6: 54.92) remains far below the opposite direction. This severe degradation suggests that Ref Qst, being more formal and structured, does not expose the model to the lexical and syntactic variability present in Raw Qst, making generalization in this direction significantly harder.
Taken together, these results reveal an important asymmetry in generalization: training on raw, naturally phrased questions provides a richer and more robust learning signal that transfers better across question styles, whereas training on cleaned reformulations leads to a model that struggles with the unpredictability of natural language input.

\newpage
\section{Models training configurations}
All models are fine-tuned using parameter-efficient LoRA adapters via the Unsloth \footnote{https://unsloth.ai} framework with 4-bit quantization and bfloat16 precision. The chat template is applied in non-thinking mode (\texttt{enable\_thinking=False}) for reasoning-based models, and training follows a supervised fine-tuning (SFT) setup using the TRL library. The key hyperparameters are summarized in Table~\ref{tab:hyperparameters}.

\begin{table}
\centering
\small
\begin{tabular}{ll}
\hline
\textbf{Hyperparameter} & \textbf{Value} \\
\hline
LoRA rank ($r$) & 32 \\
LoRA alpha & 32 \\
LoRA dropout & 0 \\
Target modules & q, k, v, o, gate, up, down proj \\
Learning rate & $2 \times 10^{-4}$ \\
Optimizer & AdamW 8-bit \\
LR scheduler & Linear \\
Per-device batch size & 4 \\
Gradient accumulation steps & 4 (effective batch size: 16) \\
Warmup steps & 5 \\
Weight decay & 0.01 \\
Number of epochs & 10 \\
Max sequence length & 2048 \\
Precision & bfloat16 \\
Quantization & 4-bit \\
Seed & 3407 \\
\hline
\end{tabular}
\caption{Fine-tuning hyperparameters.}
\label{tab:hyperparameters}
\end{table}

\end{document}